\documentstyle[FLEQN]{AUGUSTOW}
   \title{ Knowledge Acquisition, Representation \& Manipulation
   in Decision Support Systems}
\shorttitle{Knowledge Acquisition, Representation \& Manipulation}
\author{M.Michalewicz, S.T.Wierzcho\'{n}, M.A. K{\l}opotek}
\address{Institute of Computer Science\\ Polish Academy of Sciences\\       
01-237 Warsaw, ul. Ordona 21, Poland\\
e-mail: michalew{@}plearn.bitnet}

\newcommand{\oantidot}{\overline{\odot}}

\newcommand{\bdm}{\begin{displaymath}}
\newcommand{\edm}{\end{displaymath}}
\newcommand{\beq}{\begin{equation}}
\newcommand{\eeq}{\end{equation}}
\newcommand{\A}{\mbox{\bf A}}
\newcommand{\B}{\mbox{\bf B}}

\begin{document} 
\font\fui=cmfi10 % scaled \magstep1

\AugustowskiTytul

\begin{abstract}
In this paper we present a methodology and discuss some
implementation issues for a project on statistical/expert approach to
data analysis and knowledge acquisition. We discuss some general
assumptions underlying the project. Further, the requirements for a
user-friendly computer assistant are specified along with the nature
of tools aiding the researcher. Next we show some aspects of belief
network approach and Dempster-Shafer (DST) methodology introduced in
practice to system SEAD. Specifically we present the application of
DS methodology to belief revision problem. Further
a concept of an interface to 
probabilistic and DS belief networks enabling a 
 user to understand the communication with a belief network based 
reasoning system is presented
\end{abstract}

\section{Introduction}

\hspace{\parindent}A statistical database is a collection of
aggregated information at a given time. The purpose of such database
is to derive or infer facts; statistics is helpful in extracting
patterns from quantative (and qualitative) data~\cite{Chowdhury:1987}.
Data are analysed to extract information and the result of such
analysis must be validated to increase our knowledge about
modeled domain.

Several statistical packages ({\it e.g.\/} MINITAB SAS) are used
for such statistical analysis of data. However, they do not
inform the user about underlying assumptions, seldom warn against
obvious misuse, do not provide guidance in the process of
analysis, and hardly assist in the interpretation of the
results~\cite{Darius:1984}. Clearly, there is a need for statistical
expert systems \cite{French:1991}, \cite{Hand:1986},
\cite{Hand:1987}, \cite{Hand:1990}, which would include the statistical
expertise.

In \cite{Hand:1991} the authors provide a general characteristic
of statistical expert systems. They wrote:

\vspace{0.5\baselineskip}
{\narrower ``As statistical expert system is a computer program
which can act in the role of expert statistical consultant. That
is, it can give expert advice on how to design a study, what
data should be collected to answer the research question, and
how to analyse the data. Moreover, since the fundamental tool of
statistical analysis is the computer, on which the expert system
will be run, it is normally envisaged that the expert system and
data analysis routines will be an integrated whole. thus, not
only does the system advise on the analysis, but it also
actually carries it out, discussing the results and the
direction of further analyses with the user.'' \par}

\vspace{0.5\baselineskip}
In this paper we deal with some aspects of creative data
analysis. In particular we are interested in applications of AI
techniques for development of a software environment which would
allow to process both data and knowledge; such software should
provide an easy interface for formulating hypotheses and their
statistical verification.

The paper is organized as follows. The next section discusses
the architecture of a creative data analysis system. Section~3
presents knowledge acquisition strategies. Section~4 introduces
some aspects of new version of SEAD System, based on belief
network approach and DST. Section~5 is devoted to DS
methodology: after describing the nature of evidential reasoning
we present mathematics of evidential resoning and next details
of the message propagation algorithm is introduced and its
application to belief revision is presented. In Section~6 a user
interface to belief network is described.

\section{Data Analysis System}

\hspace{\parindent}Information systems are designed for storing
and manipulating data (from some domain) and for drawing
conclusions when a problem from that domain is formulated.
Clearly, this is of value insofar as the data are meaningful and
the domain formalization is sufficiently adequate.

There are at least two main categories of such systems, termed
tentatively {\it standard decision aid tools\/} (like
statistical systems, pattern recognition programs, {\it etc.\/}
and {\it knowledge based systems\/}. Although proven to be
useful, systems of both categories suffer from some drawbacks
limiting their applicability.

The main disadvantage of systems from the first category is that
a user is forced to (1) formulate his (her) problem in a formal
language, (2) check the input data validity, and (3) translate
solutions produced by the system to the language appropriate for
a given domain. In fact, to solve real-life problems the user
must cooperate with professional task analyzer.

When developing knowledge based systems an important obstacle
concerning knowledge acquisition emerges. Knowledge bases
generated by experts in different areas remain incomplete,
subjective and even contradictory. Additional difficulties arise
in cases where there is a need for processing data and knowledge
simultanously.

In both categories of systems the process of hypotheses
generation is left to the user.

The above approaches, considered separately, are especially
inefficient when formalizing poorly recognized domains or
domains with large amount of vagueness. In case of medical
diagnosis, for instance, although the characteristics of certain
diseases are fixed, the formal relationships among symptoms and
diseases are hard to determine and serious difficulties in
formulating scientific hypotheses may occur. To formalize such
domains one must combine data, partial knowledge and tentative
hypotheses.

In summary, an effective information system must provide the
following features: data storing, knowledge acquisition and
manipulation, automated generation of hypotheses, performing
standard statistical analyses, including tests of research
hypotheses, and easy communication between the knowledge base
and database.

Such experimental information system (called SEAD for
Statistical and Expert system for Analysis of Data) was recently
implemented in the Institute of Computer Science, Polish Academy
of Sciences~\cite{Michalewicz:91}.

The system architecture reflects a typical understanding of the
notion of science according which science is viewed as a
progressive development of theoretical generalizations of
observations and experiments leading to new observations and new
experiments. So, scientific research means permanent  moving
within the cycle ``observation -- generalization -- theory --
application -- observation''.

Various tools have been developed to support a researcher in
different parts of his cycle, including tools supporting storing
observations and experiments (data base systems) and storing the
acquired theoretical generalizations (knowledge base systems).
Also the tools supporting the knowledge acquisition are included,
such as the ability to generalize over a data base (statistical
data analysis systems and numerous ``learning from examples''
systems), tools allowing verification against new incoming
experiments results in the data base (statistical analysis
systems), or against new single cases (expert systems). These
tools are placed in specialized submodules, which are the main
components of the system: data base system, knowledge base
system, expert system, knowledge acquisition and verification
tools.

The main feature of this system is implementation of all four
components into one organism, whose heart is DB pumping and
absorbing data. Such a solution (based on unified data formats
and unified access methods) results in a research tool of quite
new and unique quality. The user can also employ each system's
component as an independent unit or (s)he can add his (her) own
modules.

\section{Knowledge Acquisition Strategies}

\hspace{\parindent}The completion of knowledge base by human
experts is a bottleneck in the process developing an expert
systems. Usually, experts engaged in the process are excessively
cautious (especially when dealing with poorly recognized
domains), they use different semantics (often with vague and
ambiguous notions) that results in a extremely long completion
period without any warrant that the set of rules is complete and
that it doesn't contain contradictory assertions.

Another question concerns rules validity. According to
methodological considerations, to refine a rule we search for
so-called idealizing law (being the heart of the rule) first,
and next we release systematically simplifying assumptions. Of
course, such a process must be stopped at a reasonable time, and
since some assumptions still remain hidden, the rule in its
symbolic version is valid to some extent only. Hence we may
address two questions:
\begin{itemize}
\item
were the most important assumptions embodied into the rule?, and
\item
how to estimate the degree of validity of such rule?
\end{itemize}

To answer these questions and to optimize the knowledge
acquisition process a number of tools have been developed in the
SEAD system. These tools can be divided into two categories
described in sections \ref{passive} and \ref{active}.

\subsection{Passive user}\label{passive}

\hspace{\parindent}If certain aspect of the problem under the
considerations is either ill-recognized or the knowledge about
it is too general, some data preprocessing might be necessary.
the aim of such preprocessing is to generate tentative rules
directly from an existing data base. At this stage the user
plays a passive role contrary to the further stages realized with
his active collaboration. Here the user has a few possibilities:
\begin{itemize}
\item
Discretization of a continuous  variable under a criterion of
diagnosis accuracy maximization.
\item
Some optimization but for a set of discrete variables.
\item
Identification of irrelevant variables.
\item
Generation of elementary hypotheses for specific values of variables.
\item
Generation of rules.
\end{itemize}

\subsection{Active user}\label{active}

\hspace{\parindent}By an active knowledge acquisition we mean
the simultaneous utilization of the DB and the KB, and the
statistical verification of the partial knowledge. The following
tools are helpful in these tasks.
\begin{itemize}
\item
Tools for rules identification. These the search for rules with
given condition or conclusion, the identification of rules
contexually analogical to a given rule, {\it etc.\/} Any
hypothetical rule can be therefore compared by the user with
those existing in the knowledge base.
\item
Search for data in the DB relevant to a given hypothesis;
particularly search for data contradicting given rule. these
operations are realized through queries to the DB. A specialized
interface allow simultaneous access to the KB and the DB.
\item
Rules formation from incomplete schemata (called suppositions).
This is the active strategy attempting to find a rule (or
possibly a set of rules) satisfying constraints imposed by the
user. The constraints concern values of some attributes as well
as degrees of truth which should be attached to the rules.
\item
KB verification with respect to the contents of the DB. If the
DB is sufficiently large we are interested in how accurate the
rules are, {\it i.e.\/} how strong is their expressive power
(measured in terms of misclassification error). To answer these
questions we employ the Statistical Module, which assists the
user in his research (search for the strongest relationships in
the set of variables for a given group of objects, marginal
analysis for each variable, graphics for pairs of variables,
{\it et.\/}).
\end{itemize}

\subsection{Problems of Automatic Rule Generation}

\hspace{\parindent}In very large statistical databases the
generation of hypotheses (classification rules) is done on the
basis of small samples of data. However, all known methods for
automatic generation of classification rules aim at the precise
partition of the attribute-valued space.They are create ``long''
conjunctions of the pairs (attribute, value), which precisely
classify all data sets~\cite{Quinlan:86}. Such approaches
provide reasonable results for (large) databases, where the
classification problems are well defined. However, for databases
where classification is done on the basis of small samples of
data (typical case for ``soft sciences''), this is not the case.
In such databases there is a need for preprocessing data, {\it
i.e.\/} for generating a first iteration of hypotheses which are
true with a high probability. This first iteration of hypotheses
should yield simple formulae expressed in a language
understandable for a user. The precise, ``long'' classification
rules (hypotheses) would be useless: they are precise only for a
small sample of data and don't provide a guidance for the next
iterations of hypotheses. Because of this, the development of
new classification methods for a single attribute and small
sets({\it i.e.\/} 2 or 3) of attributes are of particular
interest. Summarizing, notice, that there are several problems
connected with automatic rule generation in small databases
(these include also large databases, where a data sampling was
performed):
\begin{itemize}
\item
the set of data is not representative for drawing conclusions
strictly dividing the attribute/value space,
\item
the generated rules are hardly interpretable, if at all
possible; rules are detailed and concerned with only several
cases,
\item
a new sample (or a new set of data) can easily provide a contradiction,
\item
the set of rules usually is too large; this might be appropriate
for an expert system or diagnostic automata, but not for data
analysis,
\item
in many methods the domains of attributes must be discrete; the
process of discretization is usually left to the expert (user).
\end{itemize}

Taking into account the above drawbacks, we decided to construct
a rule generating algorithm which would start from a single
attribute (``bottom up'' approach). We present the main idea of
the method for creating the reasoning rules. This is based on
(1) an optimization process defined on the set of discrete
values of a given attribute and (2) a process of dividing a
continous domain of an attribute into subsets; such optimal
divisions are converted into the set of classification rules~\cite{MichMich}

\section{SEAD.2 -- the new version}

\hspace{\parindent}Some experiences of exploitation of SEAD
system caused several significant modifications, which were
entered into the next version of SEAD: 

\subsection{Data Base}
\hspace{\parindent}Data base, built for SEAD 1 by our group,
compatible to dBase family, appeared to be ineffective.
Especially much difficulties was noticed when cooperation of
data base with expert and statistical parts (written in C
language) was needed. We decided to accommodate CODEBASE,
which includes both libraries of procedures of
access to data bases like dBase~IV, Fox Base or CLIPPER written
in C and libraries for user interfaces near to dBase~IV or
CLIPPER for MS~DOS and MS~WINDOWS.

\subsection{Knowledge  Base}
\hspace{\parindent}Knowledge base now is the bayesian network
with uncertainty propagation based on Dempster-Schafer model, 
instead  of rule-based knowledge base with uncertainty
propagation near  to MYCIN \cite{Shortliffe:1975} used in former
version of SEAD. This changes was caused by following drawbacks
of rule approaches:
\begin{itemize}
\item
independence of variables is assumed, several appearances of the
same variable in chain causes some computation complications
(each appearance of concrete variable is treated as appearance of
separate variable)
\item
independence of rules is assumed; data base appears as a rubbish
where rules are entered de facto without any control process.
\item
Shortliffe's approach assumed the use certainty factor CF
\cite{Shortliffe:1975}, nature of which was defined not clearly
enough. Even if rule's CF was interpreted in terms of
probability, combination of rules lead very fast to loss of
interpretability.
\end{itemize}

Meanwhile the belief networks has the following advantages:
\begin{itemize}
\item
there exists well defined relationship between variables
(conditional or not)
\item
these relationships are to be tested very simply, what can be
used for new rules generation for example
\item
one cannot add/remove rules without retribution, because it can
break relationships between variables
\end{itemize}

Approach based on belief networks and Dempster-Schafer
methodology generates many theoretical and practical problems.
Some of them are showed below in section \ref{piata}

\subsection{Rule preceiving of bayesian knowledge base}

\hspace{\parindent}It is easy to notice, that rule approach has
several significant advantages. The most important of them are
following: simplicity, rule interpretability and naturalness and
comparatively easiness of introducing some knowledge acquisition
mechanisms.

That was the reason of introducing to new version of SEAD
systems some special methods, which make possible for the user
to preceive bayesian knowledge base as a set of rules. These
methods are described below in section \ref{szosta}.  

\section{DS Approach to Knowledge Representation and Manipulation}
\label{piata}

\subsection{The Nature of Evidential Reasoning}

     The term {\it evidential reasoning\/} coined by 
Lowrance and Garvey in 
\cite{Lowrance:1982} 
covers a set of techniques  designed  for  processing  and 
reasoning from row data ({\it evidence\/}) which are currently accessible 
to an agent. It is based upon the Dempster-Shafer theory (DST) of 
belief functions. This theory was founded by  G.  Shafer  in  his 
monograph \cite{Shafer:76} and was highly
 influenced by the works  of  A.  P. 
Dempster \cite{Dempster:1966,Dempster:1967}. 
DST is based on two principal  ideas:  ``the 
idea of  obtaining  degrees  of  belief  for  one  question  from 
subjective probabilities for a related  question  and  Dempster's 
rule for combining such degrees of belief when they are based  on 
independent items of evidence'' \cite{Shafer:1990} p. 474. 

     To give a concise rationale for this theory 
(see \cite{Ruspini:1992})  for  deeper  discussion)  
we  should  note  that  the 
information required to  understand  the  current  state  of  the 
world, or to solve a real problem, comes from  multiple  sources, 
like real-time sensor  data,  accumulated  domain  knowledge  and 
current contextual information. Information coming from  sensors, 
called {\it evidence\/}, provides a support for certain conclusions.  The 
very nature of evidence can be summarized as  follows:(1)  it  is 
{\it uncertain\/}, i.e it allows for multiple possible explanations,  (2) 
it is {\it incomplete\/} as  the  sensors  rarely  has  a  full  view  of 
situation, (3) it may be completely or partial {\it incorrect\/}.

     Because of its nature, evidence is not  readily  represented 
either by logical or by classical probabilistic formalism. The {\it ad 
hoc\/} proposed methods for  handling  uncertain  information  (like 
e.g. CF factors) cause  difficulties  in  interpretation  and  in 
extending  the  capabilities  of  such  methodologies.  DST,   in 
contrast,  provides  a  natural  representation  for   evidential 
information,  a  formal  basis  for  drawing   conclusions   from 
evidence, and a representation for belief.

     In evidential reasoning a knowledge  source  is  allowed  to 
express probabilistic  opinions  about  the  (partial)  truth  or 
falsity of statements composed of subsets of propositions from  a 
space of distinct, exhaustive possibilities (called the {\it frame  of 
discernment\/}). The beliefs  can  be  assigned  to  the  individual 
(elementary) propositions in the  space  or  to  disjunctions  of 
these propositions (i.e. to compound propositions) or both.  When 
belief is assigned  to  a  disjunction,  a  knowledge  source  is 
explicitly stating that it does not  has  enough  information  to 
distribute this belief more precisely. Hence a  knowledge  source 
can distribute its belief  to  statements  whose  granularity  is 
appropriate to its state of knowledge. Further, the statements to 
which belief is assigned are not necessarily  distinct  from  one 
another. The distribution of beliefs over a frame of  discernment 
is called a {\it body of evidence}.

     The easiest way to formally represent a body of evidence  is 
to define a mass function m being a probability distribution over 
a set of subsets of a frame of discernment $\Theta$  (as  such  it  must 
satisfy the obvious condition that 
$\sum \{m(A)\mid A \subseteq \Theta\} = 1$).  Now,  if 
the mass function is positive on singletons of $\Theta$  only  then  our 
model reduces to standard probability distribution. On the  other 
extreme, a mass function can take the form $m(A) = 1$ for some 
$A \subseteq \Theta$. 
When $A = \Theta$ then such a  function  represents  total  ignorance 
about current state of the world. When  $A \subset \Theta$,  then  the  mass 
function describes our certainty that the truth lies in $A$; when $\Theta$ 
is a Cartesian product of  frames  describing  related  questions 
then $m(A) = 1$  defines a logical belief  function.  Thus,  {\it Boolean 
logic\/} is another special case of belief functions.

     DST is therefore a generalization of both Boolean logic  and 
probability  calculus.  This  generalization  lies  deep  in  the 
representation of  belief  functions  and  has  effects  on  many 
aspects of the theory. The heart of  this  theory  is  {\it Dempster's 
Rule of Combination\/}, or DRC for short, being a  formal  tool  for 
fusing different bodies of evidence. The result is a new body  of 
evidence representing the consensus among the original bodies  of 
evidence. In special cases this rule behaves like Bayes' rule and 
in other  cases  like  set  intersection.  Thus  belief  function 
provide  a   framework   in   which   propositional   logic   and 
probabilistic statements can be freely mixed and processed  under 
common formalism.

     Belief functions,  however,  provide  more  general  models. 
Particularly, random sets are a generalization of the fixed  sets 
of logical models and upper {\it (plausibility  functions)\/}  
and  lower {\it (belief functions)\/} bound intervals are a 
generalization of  point probabilities.

     Current automated reasoning systems are most effective  when 
domain  knowledge  can  be  modeled   as   a   set   of   loosely 
interconnected  concepts   (propositions)   what   justifies   an 
incremental approach to updating beliefs. Applying DST formalism, 
independent opinions  are  expressed  by  independent  bodies  of 
evidence and dependent opinions can  either  be  expressed  by  a 
single  body  of  evidence  or  by  a  network   describing   the 
interrelationships among several bodies of  evidence  (so  called 
{\it belief networks\/}). 
Updating the belief (by using  Dempster's  Rule 
of combination) in one proposition affects  the  entire  body  of 
evidence. This allows to  involve  whole  the  knowledge  already 
accumulated in the system (contrary to rule-based  systems  where 
only fragmentary knowledge is used during the process).

     In this paper we  assume  some  familiarity  with  the  DST, 
although the appropriate definitions are  included  in  the  next 
section. In the sequel  we  will  focus  on  the  mathematics  of 
evidential reasoning, mainly on so-called belief revision.

\subsection{Mathematics of Evidential Reasoning}

     In  this  section   we   briefly   review   a   mathematical 
formalization of the evidential reasoning.  It  consists  of  two 
main parts. In static part  we  simply  represent  our  knowledge 
about a problem in terms of belief  functions  and  dynamic  part 
allows to express knowledge evolution.

\subsubsection{Basic notions \protect \\}

     ${\cal X} = \{X_1 ,\ldots ,X_n \}$ 
  is a set of variables and $\Theta_i  = Dom(X_i )$ is  a 
discrete set of possible  values  of  $i$-th  variable.  (The  case 
of  continuous  variable  is  considered  by  \cite{Shafer:1979}  or 
\cite{Wasserman:1990}.)

     By {\bf A}, {\bf B}, {\bf C},\ldots we shall denote 
  subsets of the set ${\cal X}.\Theta$  will 
denote the configurations space or  frame  of  discernment,  
$\Theta  = \times \{\Theta \mid i = 1\ldots n\}$, and 
$\Theta.\mbox{\bf A} = \prod\{\Theta \mid X_i  \in \mbox{\bf A}\}$ 
will denote  the  marginal 
frame. If {\bf x} is a configuration from $\Theta$  then  {\bf x.A}  
stands  for  the 
projection of {\bf x} onto $\Theta.\mbox{\bf A}$; 
{\bf x.A} is obtained by dropping from {\bf x}  the 
components being members of $\Theta_i$  for all 
$X_i\in {\cal X}\setminus \mbox{\bf A}$.

\subsubsection{Knowledge representation\protect\\}

     Assume $(\Omega,{\cal B},P)$ is a probability space, 
     $\Xi$ is a discrete space 
and $X$ is a mapping which assigns subsets of $\Xi$  to  points  of  
$\Omega$, 
i.e. $X(\omega)\subseteq\Xi$ for each $\omega \in \Omega$. 
Treating $\Xi$ as the set of elementary 
hypotheses we identify each subset of $\Xi$ as a compound hypothesis. 
If $\Omega$ is viewed as a set of observations then  $X(\omega)  =  A$  
can  be 
interpreted as a rule ``If observation = $\omega$ then hypothesis  =  $A$''. 
According to classical approach the mapping $X$ induces a measure $\mu$ 
defined on POW - the  power  set  of  the  set  of  all  compound 
hypotheses:

\beq
     \mu( A) = P(\{\omega\in \Omega\mid X(\omega) \in  A \}), 
     \quad  A \in \mbox{POW} 
\eeq
Note that when $X$ is a function, i.e. 
$X(\omega)\in \Xi$ for all $\omega\in \Omega$  then 
POW can be reduced to the power set of $\Xi$ and (1) reduces  to  the 
definition of the random variable. Under the  general  setting  $X$ 
constitutes so-called random set~\cite{Wierzchon:1993a}.

     Of special importance are the  next  set  functions  derived 
from (1) and defined for all $A\subseteq \Xi$, $A\neq\emptyset$:
\beq
     m(A)   = P(\{\omega\in \Omega\mid X(\omega) = A\}), 
\eeq
\beq    
     Bel(A) = P(\{\omega\in \Omega\mid X(\omega)\subseteq A, \ 
     X(\omega) \neq \emptyset\}) 
\eeq
\beq
 Pl(A)  = P(\{\omega\in \Omega\mid (X(\omega) \cap A) \neq \emptyset\}),
\eeq
Assuming that all observations are relevant, i.e. 
$X(\omega)  \neq  \emptyset$  for 
all $\omega\in \Omega$, we easily recover the main  properties  of  the  basic 
probability assignment, m, i.e. $m(\emptyset) = 0$,
and $\sum \{m(A)\mid A\subseteq \Xi\} =  1$. 
In general case, when $m(\emptyset) >  0$,  the  mass  function  should  be 
modified according to the assignment  $m(A)  :=  m(A)/(1-m(\emptyset)$  and 
$m(\emptyset) = 0$.  The  $m$-function  plays  the  role  analogical  to  the 
probability distribution $P(X=x)$, $x\in \Theta_X$; particularly  when  
$m(A)$ 
is positive on the singletons only then $m$ is just  a  probability 
distribution.  The  belief  and  plausibility  functions  can  be 
computed from $m$ by means of the identities

\beq
  Bel(A) \, = \, \sum \{m(B)\mid B\subseteq A\}, \ 
  Pl(A) = \sum \{m(B)\mid B \cap A \neq \emptyset\}
  \eeq
Similarly plausibility function can  be  deduced  from  a  belief 
function by means of the equation $Pl(A) = 1 - Bel(\Xi - A)$, $A\in \Xi$.

 Shafer \cite{Shafer:76} proposes further formulas enabling to  express 
one set  function  in  terms  of  other  set  function  and  fast 
algorithms for doing this task are reported in \cite{Kennes:1990}.

     In the sequel we shall use  multivariate  belief  functions, 
i.e. belief function defined on the  family  of  subsets  of  the 
space $\Theta$ introduced at the beginning of this section.

     Special class  of  multivariate  belief  functions  form  so 
called {\it logical\/} belief functions 
satisfying the condition $m(A) = 1$ 
for some $A \subset \Theta$. 
(When $A = \Theta$ then  the  mass  function  represents 
{\it vacuous\/} belief function; it  models  total  ignorance  about  the 
possible location of ${\cal X} in \Theta$.) 
To be more illustrative we  present 
four elementary situations:
\begin{itemize}
\item[(a)]
 the logical expression ``$X_1  = x_1 \ \mbox{and\ } 
 X_2  = x_2 \ldots \ \mbox{and\ } X_n   =  x_n$'' 
is represented by the element $\theta = <x_1 ,x_2 ,\ldots ,x >\in \Theta$,
\item[(b)]
 the logical expression 
 ``$X_1  = x_1 \ \mbox{or\ } 
 X_2  = x_2 \ldots \ \mbox{or\ } X_n   =  x_n$''is 
 represented by the subset of the form  
 $\bigcup_{i=1}   ^n (\{x_i \}\times \Theta.({\cal X}-\{X_i \}))$,
\item[(c)]
 the logical expression ``not~$(X_1  = x_1)$'' is represented by  the 
 subset of the form $(\Theta_1 -\{x_1 \}) \times \Theta.({\cal X}-\{X_1\})$.
\item[(d)]
 the logical expression ``if $((X_1  = x_1 )\ \mbox{or\ }\ldots\,\mbox{or\ } 
  (X_{n-1} = x_{n-1}) \quad \mbox{then \ }
   (X_n  = x_n )$'' is represented by  the  subset  of  the  form 
$\{x_1 ,x_2 ,\ldots ,x_n\} \cup 
(\Theta_1 -\{x_1\})\times(\Theta_2 -\{x_2 \})\times 
\ldots \times (\Theta_{n-1}-\{x_{n-1}\})\times\Theta_n$ .
\end{itemize}

     When the logical  assertions  cannot  be  stated  with  full 
certainty we convert them to so called simple support function by 
setting $m(A) = \alpha$ and 
$m(\Theta) = 1-\alpha$, where $\alpha \in  (0,1)$  and  $A$  is  a 
subset of $\Theta$ representing appropriate assertion.

\subsubsection{Knowledge evolution\protect \\}

     Up to this moment  we  have  assumed  that  all  the  belief 
functions are defined on the common space $\Theta$. Suppose we have  two 
belief functions defined on the frames $\Theta\mbox{\bf .A}$ 
and $\Theta\mbox{\bf .B}$, respectively. 
How to represent beliefs expressed by  means  of  the  attributes 
(variables) from the set {\bf A} in terms of the  attributes  from  the 
set {\bf B}? We answer these questions now.

    {\it Minimal extension\/}. 
    Suppose that to describe a domain we used 
attributes from the set $\A \subset {\cal X}$ 
and as a result we obtained  belief 
function $Bel_{\A}$  defined over 
the space $\Theta\mbox{\bf .A}$. 
After a time we decided 
to use larger set of attributes $\B \supset \A$. 
To translate the  function 
$Bel_{\A}$  into the belief function $Bel_{\B}$  defined  over  the  space  
$\Theta\mbox{\bf .B}$ 
observe that using the attributes from the set {\bf A} we are  able  to 
create $m = \prod_{i\in\A }\mid\!\!\Theta_i\!\!\mid$ 
(where $\mid\!\!\Theta_i\!\!\mid$ stands for the cardinality  of  the 
set $\Theta_i$ ) elementary propositions q of the form 
``$X_{i_1}   = x_{i_1}$   and  
$X_{i_2}   = x_{i_2}$   and \ldots and 
$X_{i_r}   = x_{i_r}$'' where $X_{i_j}  \in \A$, $x_{i_j}  \in \Theta_{i_j}$
   and  $\mid\!\!\A\!\!\mid = r$. 
In other words the $\theta$'s are the only atomic propositions of the 
language $L_{\A}$. Since the values of the attributes form the set  $\B-\A$ 
are unknown to a subject using the language $L_{\A}$  then  each  atomic 
proposition $\theta\in \Theta\mbox{\bf .A}$  translates  into  
the  compound  proposition 
$\theta\times\Theta\mbox{.}(\B-\A) \subset \Theta\mbox{\bf .B}$. 
Similarly each  $A \subseteq  \Theta\mbox{\bf .A}$  translates  into  the 
subset  $A \times  \Theta\mbox{\bf .}(\B-\A)$  of  $\Theta\mbox{\bf .B}$.  
Thus  the   mass   function   $m_{\A}$  
characterizing the belief function $Bel_{\A}$  translates into the  mass 
function $m_{\B}$  of the form

\beq
     m_{\B} (B) = \left\{\begin{array}{ll}
m_{\A} (a) & \mbox{\ if \ } B=A\times\Theta\mbox{\bf .}(\B - \A), \ 
A\subseteq \theta\mbox{\bf .A} \\
0 & \mbox{\ otherwise}
     \end{array}
     \right.                                                 
\eeq
In the sequel we shall denote the function $m_{\B}$  derived from $m_{\A}$
 by means of the above procedure as $m_{\A}   ^{\uparrow{\B}}$.

  {\it Marginalization\/}. Assume now that we  reduce the set {\bf B}
    of 
attributes to its subset {\bf A}. Obviously each atomic proposition 
$\theta \in \Theta\mbox{\bf .B}$ points to a proposition 
$\xi \in \Theta\mbox{\bf .A}$ if its  projection  onto  
$\Theta\mbox{\bf .A}$ 
(defined as ``dropping coordinates form $\B-\A$'') equals  to  
$\xi$.  Thus 
the mass function $m_{\B}$  translates to a mass $m_{\A}$  via the equation
\beq
     m_{\A} (A) = \sum \{m_{\B} (B)\mid 
     B \subseteq \Theta\mbox{\bf .B}, \ proj(B) = A\}
\eeq
In the sequel we shall denote the function $m_{\A}$  derived from 
$m_{\B}$   by 
means of the above procedure as $m_{\B}   ^{\downarrow\A}$   .

{\it Fusing evidence\/}.  When  $m_1$   and $m_2$ are  two  separate  and 
independent bodies of evidence defined on a common  frame  $\Theta$,  to 
arrive to the final conclusions  we  combine  them  by  means  of 
{\it Dempster's Rule of Combination\/} (DRC for short). That is we  crate 
new mass function 

\beq
\begin{array}{lcl}
     m(A) & = & (m_1 \oplus m_2 )(A) \\
&= &
\kappa \cdot\sum
 \{m_1 (A \cup  B)\cdot m_1 (A \cup C)\mid \,
  A,B \subseteq (\Theta-A), A\cap B = \emptyset\}
\end{array}
\eeq
where $\kappa$ is a normalizing constants defined as 
$\kappa   ^{-1}   = 1 - m(\emptyset)$.

     Note that when $m_1$  and $m_2$   are  categorical  mass  functions, 
i.e. $m_1 (A) = 1$ and $m_1 (B) = 1$  for some  $A,B \subseteq  \Theta$  
then  the  rule 
produces categorical mass function with $m(A\cap B)  =  1$.  Hence  DRC 
generalizes classical sets intersection.

     To illustrate the notions introduced in this section  assume 
that $p$ and $q$ are two propositional variables, 
and their frames  $\Xi$ 
and $\Lambda$ are  such  that  
$\Xi  =  \Lambda  =  \{t,f\}$  where  $t$  and  $f$  stand 
respectively 
for ``true'' and ``false''. The logical implication can be  described 
by the mass function $m_{p\rightarrow q}$    on 
$\Theta = \Xi\times\Lambda$  with  one  focal  (i.e. 
positive) element $m_{p\rightarrow q}
(\{\{f,f\}, \{f,t\}, \{t,t\}\}) = 1$. Similarly  the 
fact ``$p$ is true'' is represented by the categorical mass  function 
$m_p$  on the frame $\Xi$, that is $m_p (\{t\}) =  1$.  
To  combine  these  two 
functions we must extend $m_p$  to the bpa on 
$\Theta$: $m_p   ^{\uparrow\Theta}  (\{t,f\}, \{t,t\}\}) = 
1$.  Now, combining $m_{p\rightarrow q}$    with 
$m_p   ^{\uparrow\Theta}$   we obtain the bpa $m$ of  the  form 
$m(\{t,t\}) = 1$. Projecting this $m$ onto the space $\Lambda$  we  
state  that 
$m   ^{\downarrow\Lambda}(\{t\}) = 1$ what means that 
``$q$ is true''  what  agrees  with  the 
standard logical inference pattern {\it Modus Ponens\/}. 
Note that if  we 
combine $m_{p\rightarrow q}$    with $m$-function representing 
statement ``$p$ is  false'' 
then the resulting function (after projection on the frame $\Lambda$) has 
the form $m(\{\{f\}, \{t\}\})$, i.e. nothing can be said about the  truth 
of the proposition $q$ (what again agrees  with  classical  logic). 
Similarly combining $m_{p\rightarrow q}$
    with the $m$ function stating that  ``$q$  is 
false'' we obtain a counterpart of the {\it Modus Tollens\/} pattern.

     This example shows that the formalism  of  belief  functions 
perfectly agrees with the  propositional  calculus.  In general, 
reasoning with categorical belief functions  corresponds  to  the 
idea of using set valued description of variables  to  estimation 
and testing hypotheses - cf \cite{Schweppe:1973}.

{\it Conditioning\/}. When $Bel$ is a belief function over 
$\Theta$ and  $Bel_o$  
is a categorical belief function focused on a subset $A$ of $\Theta$  then 
the combination of $Bel$  with  $Bel_o$   results  in  the  conditional 
belief  function  $Bel(\cdot\mid A)$.  Its  properties  are  described   in 
\cite{Shafer:76}, \cite{Wierzchon:1987} and \cite{Smets:1988}.

\subsection{Reasoning under Uncertainty}

     With the DRC the problem  of  reasoning  can  be  stated  as 
follows: Given a  collection  of  unrelated  pieces  of  evidence 
$\varepsilon_1 ,\varepsilon_2 ,\ldots ,\varepsilon_m$  
(each of which  is  translated  to  the  appropriate 
belief function) we turn them into a single body of the form 
$\varepsilon_1\wedge \varepsilon_2\wedge \cdots\wedge \varepsilon_m$ 
characterized  by  a  belief  function  being  the 
orthogonal sum the appropriate belief functions.  Adding  to  the 
resulting body some  findings  which  are  possibly  certain,  we 
obtain a conditional belief function. This closely corresponds to 
the probabilistic reasoning where  the  bodies  of  evidence  are 
represented by the (conditional)  probabilities  from  which  the 
joint  probability   distribution   is   derived;   having   this 
distribution we are looking for a distribution conditioned  on  a 
subset of $\Theta$.

 To formalize the  problem  observe  that  each  body  $\varepsilon_i$  is 
represented as a belief function over a subset $H_i \subseteq {\cal X}$. 
Denoting ${\cal H}$ 
the family of subsets $H_i$, $i = 1\ldots m$ we are interested in a  belief 
function
\beq
     Bel_X (\cdot\mid \mbox{\bf e}) = (\oplus \{Bel | H\in {\cal H}\})
        ^{\downarrow\{X\}} 
\eeq
where {\bf e} is a subset of  $\Theta$  representing current findings.  This 
corresponds to deriving a  conditional  probability  distribution 
$P(X\mid \mbox{\bf e})$ from the joint probability 
distribution $P(X_1 ,\ldots ,X_m )  =  \prod 
\{P(X_i \mid\Pi(X_i )\mid i = 1\ldots m\}$ 
given the set of observations (instantiated 
variables) {\bf e}; here $\Pi(X_i )$ is the set of immediate 
causes for $X_i$   -- 
see \cite{Pearl:1}. Note  that  under  probabilistic  setting  the 
subsets $H_i$  are defined as the sum of $\{X_i \}$ 
and $\Pi(X_i )$; further  the 
operator $\oplus$ is replaced by the multiplication and  the  projection 
operator $   ^{\downarrow\{X\}}$
corresponds to the  summation  over  all  variables 
form the set ${\cal X}-\{X\}$. Hence the equation (9) nicely summarizes both 
the probabilistic and strictly evidential problem of reasoning.

     Because immediate derivation  of  $Bel_X (\cdot\mid\mbox{\bf e})$ 
      is  numerically 
very expensive, a number of procedures has been designed to  this 
problem - cf. \cite{Pearl:1}, \cite{Lauritzen:1988},
\cite{Shenoy:90}.

     The simplest and more attractive procedure seems to be  that 
of Shafer  and  Shenoy.  It  relies  upon  the  observation  that 
deleting a single variable $Y$ from the set ${\cal X}$ 
results in  a  belief 
function (here ${\cal X}' = {\cal X} - \{Y\})$
\bdm
     Bel    ^{\downarrow{\cal X}'} 
         = (\oplus\{Bel_H \mid 
         H\in {\cal H}, \, Y \not\in H\}) 
         \oplus (\oplus{Bel_H \mid H\in {\cal H}, Y\in H}) 
            ^{\downarrow{\cal X}'} 
\edm
and follows from the fact that (1) the operator $\oplus$ is  commutative 
and associative, and (2) if $Bel_{\A}$  and $Bel_{\B}$  
are two belief function 
defined over $\Theta\mbox{\bf .A}$ and 
$\Theta\mbox{\bf .B}$, respectively, then 
$(Bel_{\A}\oplus  Bel_{\B})    ^{\downarrow\A}   = 
Bel_{\A}\oplus  Bel_{\B}   ^{\downarrow(\A\cap\B)}$.
Note  that  after   performing   the   above 
computations the original family  ${\cal H}$  reduces  to  new  
family  ${\cal H}'$ 
consisting of all $H$'s such that $Y \not\in H$, and the new set  
$H'  =  (\bigcup 
\{H\mid  Y\in H\}) - \{Y\}$. In its very nature the  procedure  is  closely 
related to the approach used by \cite{Bertelr:1972}  in 
solving nonserial dynamic problems and it says  that  to  find  a 
margin over $\Theta.\{X\}$ it suffices to delete one by one variables from 
the  set  ${\cal X}-\{X\}$,  each  time  combining  belief  functions   over 
relatively small frame being the sum of these $H$'s  which  contain 
the currently removed variable. This last  operation  is  allowed 
because the projection operator is such that (3) if $I \subset J$
 are two 
subsets of ${\cal X}$ then 
$Bel   ^{\downarrow I}= (Bel   ^{\downarrow J})   ^{\downarrow I}$. 
The  three  properties  (1), 
(2) and (3)  can  be  viewed  as  the  necessary  and  sufficient 
conditions for local computations described  below-  cf. 
\cite{Shenoy:90}.

     Obviously, the numerical complexity of  this  new  procedure 
hardly depends on the order in which the variables from  the  set 
${\cal X}'$  are removed - see \cite{Bertelr:1972} for  details. 
Shafer and Shenoy (1986) observed that an optimal  order  can  be 
recovered if we arrange the original family ${\cal H}$ into a 
join tree  -- 
consult \cite{Maier:1983} -- (Markov tree in their nomenclature) which 
has a nice property stating that when $L\in {\cal H}$ 
is a leaf node in the 
tree and $P\in {\cal H}$ is its parent node then the variables from the set 
$L-P$ occurs only in the set $L$. (Of course when ${\cal H}$ is not an acyclic 
hypergraph it must be embedded in such a hypergraph first  --  see 
\cite{Wierzchon:1993b} for an appropriate algorithm).

     Now the procedure of finding $Bel   ^{\downarrow {\cal X}'}$
          can  be  summarized  as 
follows. Assume that ${\cal H}$ is organized into a join  tree  rooted  on 
the node corresponding to the variable $X$. Each node in  the  tree 
can be imagined as  a  separate  processor  equipped  with  local 
information in the form of a component belief  function  (when  a 
node was added to ${\cal H}$ in the process  of  the  tree  formation  its 
local information has the form of vacuous belief function).  Each 
node communicates with its neighbors by passing messages  (belief 
functions) to the neighbors. There are two kinds op operations: a 
{\it propagation\/} operation which describes  how  messages  are  passed 
from node to node  so  that  all  of  the  local  information  is 
globally distributed, and a {\it fusion\/} rule which describes  how  the 
messages incoming to a node are combined to make marginal  belief 
functions and outgoing messages. In other words propagation takes 
place along the edges of the tree and fusion takes  place  within 
the nodes. Both the operations can be formalized as follows.  Let 
$N(H;{\cal T})$ denote the set of all neighbors of the node 
$H$ in the  join tree ${\cal T}$. 
The message which the node  $H$  sends  to  a  neighbor  $H'$ 
equals to
\beq
     M   ^{H\rightarrow H'}
       = (Bel_H \oplus (\oplus \{M   ^{G\rightarrow H}
       \mid G\in (N(H;{\cal T})-\{H'\})\}))   ^{\downarrow(H\cap H')}   
\eeq
The operation  in  the  parentheses  corresponds  to  fusion  and 
$(\cdot)   ^{\downarrow(H\cap H')}$
represents propagation.  Note  that  (10)  allows  for 
parallel belief updating. When a node received messages from  all 
its neighbors, a marginal belief is computed
\beq
     Bel   ^{\downarrow H}
        = Bel_H  \oplus (\oplus \{M   ^{H'\rightarrow H}
            \mid H'\in N(H;{\cal T})\})          
\eeq
This procedure with some modifications  reducing  the  number  of 
summations  $\oplus$,  described  in  \cite{Wierzchon:1993a}.,   has   been 
successfully implemented in our system.

\subsection{Belief Revision}

     The aim of belief revision is, according  to 
\cite{Pearl:1} Ch. 5
to identify a composite set of propositions (one from each 
variable) which ``best''  explains  the  evidence  at  hand.  Under 
probabilistic context this formalizes as follows.  Given  {\bf e},  the 
set of instantiated variables (evidence),  let  {\fui x}  stand  for  an 
assignment of values to the variables in  ${\cal X}$  consistent  with  {\bf e}; 
such an {\fui x} is said to be {\it explanation\/}, 
{\it interpretation\/} or  {\it extension\/} 
of e. The {\it most probable explanation\/} (MPE) of the evidence at hand 
is  such  an  extension  $\mbox{\fui x}_{\mbox{\bf e}}$   
which  maximizes   the   conditional 
probability $P({\cal X} =\mbox{\fui x}\mid \mbox{\bf e})$. 
This  task  can  be  performed  locally  by 
letting each variable $X$ in ${\cal X}$ compute the function
\beq
     \beta(x\mid\mbox{\bf e}) = \max_{\mbox{\fui x'}} \{P(x,\mbox{\fui x'})
     \mid \mbox{\bf e}\}
     \label{dwana} 
\eeq
where {\fui x'}  stands  for  an  explanation  projected  on  the  space 
$\Theta({\cal X}-\{X\})$. That \ref{dwana}  
can be computed locally follows from the fact 
that  the  pair  $(\cdot ,\max)$  satisfies  the   requirements   (1)-(3) 
specified in section 5.3.  Note  that  in  \ref{dwana}  the  operation  of 
summation -- requested when  computing  conditional  probabilities 
$P(X\mid \mbox{\bf e})$ -- was replaced by the maximum operator.

     When we use belief function  the  maximum  operator  appears 
both in the $\oplus$-summation and in the projection operator. Hence  to 
fit the rule (9) to the formula (12) we must  redefine  both  the 
operators. First, instead $\oplus$-summation we must  define  new  
$\oplus-max$ operator
\beq
     (m_1  \oplus_{max}m_2 )(A) = 
     \max\{m_1 (A\cup C)\cdot m_2 (A\cup D)
     \mid C,D\subseteq (\Theta-A), \ C\cap D = \emptyset\}  
 \eeq
Here, contrary to the $\oplus$-summation, normalization  is  unnecessary 
as the relative strengths are more important  than  the  absolute 
values of commitment to a given set of  propositions.  Similarly, 
the projection operator takes now the form (we assume that 
$\A \subset \B$) 
\beq
     m_{\A}   ^{\downarrow max\B}
   = \max \{m_{\B} (B)\mid  B\subseteq \Theta.\B, proj(B) = A\}
\eeq
Using these operators  in  equation  (10)  we  obtain  a  message 
passing algorithm for belief updating. This algorithm can be used 
in two main modes described below.

{\it Explanatory mode\/}. Rooting the join tree on a variable  $X$  we 
are searching for the instantiations  of  other  variables  which 
explain computed value ox $X$ at best.

{\it Hypothesizing\/}. Again we root the tree on a variable $X$ and we 
assume that $X$ takes a value $x_o$. Our algorithm allow to  find  the 
best explanation for such an assignment (i.e. we are searching an 
answer  for  the  question  ``which  settings  of  the   remaining 
variables explain the condition ``$X = x_o$'' at best?'')

  {\it Conditioning\/}. 
  We choose a group of variables, say $\varepsilon$, and  we 
instantiate them to the values $\mbox{\bf e}\in  \Theta.\varepsilon$. 
Now our algorithm  allows 
to answer the question ``what will happen if $\varepsilon = \mbox{\bf e}$?''.

\section{Viewing Belief Network as a Set of Rules}
\label{szosta}
A natural phenomenon of human expression of knowledge is a widespread usage 
of implications: most mathematical theorems are of the form "if $<premise>$ 
then $<conclusion>$", though obviously they are equivalent to for example a 
 CNF or DNF representation. Furthermore, humans use implication-like 
statements 
even if described phenomena fail to be deterministic, e.g. in describing 
 course of chemical reactions, laws in social sciences etc. The term "rule" 
for implication-like expression of knowledge found a wide-spread use in 
artificial intelligence, e.g. in MYCIN-like expert systems. %\\

Taking into account this phenomenon, also our group made an effort to find a 
way for describing the contents of a belief-network based knowledge bases 
(which is very different in nature from modular production rules based 
knowledge bases) in terms of rule-like constructs. %\\

This transcription should support the user by helping him:
\begin{itemize}
\item to understand the contents of the knowledge base%\\
\item to formulate queries to the knowledge base%\\
\item to enter and update the knowledge base and %\\
\item to understand the justifications of the system for results of its 
reasoning%\\
\end{itemize}

The last item has just been subject of the preceding sections.

To understand better the the nature of the rule-like transcription
let us first give the general definition of a belief network:%\\

\subsection{The Concept of a Belief Network}

We generalize here the definition of belief network from \cite{Geiger:90}
(bayesian networks), while using the denotation of 
\cite{Shenoy:90}.  %\\
\begin{definition}
We define               a mapping $\oantidot: VV \times VV 
\rightarrow VV$ called decombination such that: \\
if $BEL_{12}=BEL_1 \oantidot BEL_2$ then $BEL_1=BEL_2 \odot BEL_{12}$.\\
 \end{definition}%\\
In case of probabilities,
 decombination means memberwise division: 
$$Pr_{12}(A)=Pr_1(A)/Pr_2(A).$$ 
In case of DS belief functions, 
let us remaind that 
                the commonality function Q is defined as
$Q(A)=\sum_{B;A \subseteq B} m(B)$. Hence  for  $Bel_{12}=Bel_1 \oplus 
Bel_2$ then $Q_{12}(A)=c \cdot Q_1(A) \cdot Q_2(A)$ (c - normalizing factor).
Let us remaind also that by a pseudo-belief function a function over powerset 
is understood which differs from the proper belief function by the fact, that 
also negative mass function (m) values  are allowed, but only to such extent
that the commonality function Q remains non-negative.
Then let us define the operator $\ominus$ as yieldiing  a DS pseudo-belief 
function such that: whenever $Bel_{12}=Bel_1 \ominus Bel_2$ 
 then $$Q_{12}(A)=c \cdot Q_1(A)/Q_2(A)$$ (c - normalizing constant) for 
non-zero $Q_2(A)$. It is easy to 
check that such a function always exists. Obviously,  $Bel_{1}=Bel_1 \oplus 
Bel_{12}$. The operator $\ominus$ means then the decombination of two DS 
belief functions. 
 Both for probabilities and for DS belief 
functions decombination may be not uniquely determined. Moreover, for DS 
belief functions not always a decombined DS belief function will exist. Hence 
we extend the domain to DS pseudo-belief functions which is closed under this 
operator. We claim here without a proof (which is simple) that DS 
pseudo-belief 
functions fit the axiomatic framework of Shenoy/Shafer. Also, we claim 
that if an (ordinary) DS  belief  function  is  represented  by  a 
factorization in 
DS pseudo-belief functions, then any propagation of uncertainty yields the 
very 
same results as when it would have been factorized into ordinary DS belief 
functions. Let us define now the concept of pseudo-conditioning.%\\
\begin{definition}
By pseudo-conditioning $|$ of a belief function $BEL$ on a set of variables 
$h$ 
we understand the transformation: $BEL ^{|h}= BEL \oantidot BEL ^{\downarrow 
h}$. %\\
\end{definition}%\\
Notably, pseudo-conditioning means in case of probability functions proper 
conditioning. In case of DS pseudo-belief functions the operator $|$ has 
meaning entirely different from traditionally used notion of conditionality 
- pseudo-conditioning is a technical term used 
exclusively 
for valuation of nodes in belief networks. Notice: some other authors 
e.g. \cite{Cano:93} recognized also the necessity of introduction of two 
different notions in the context of the Shenoy/Shafer axiomatic framework 
(compare a priori and a posteriori conditionals in \cite{Cano:93}). 
\cite{Cano:93} introduces 3 additional axioms governing the 'a priori' 
conditionality to enable propagation with them.  
Our 
pseudo-conditionality is bound only to the assumption of executability of the 
$\oantidot$ operation and does not assume any further properties of it. 
We will discuss the consequences of this difference elsewhere.
Let 
us define now the general notion of belief networks:%\\
\begin{definition}
 A 
belief 
 network is a pair (D,BEL) where D is a dag (directed acyclic graph)
and BEL  is a belief 
distribution called the {\em underlying distribution}. Each node i in D 
corresponds to a variable $X_i$  in BEL, a set of nodes I corresponds to a 
set of variables $X_I$ and $x_i, x_I$
 denote values drawn from the domain of $X_i$ 
 and from the (cross product) domain of $X_I$ respectively. Each node in the 
network  is regarded as a storage cell for any  distribution 
$BEL ^{\downarrow \{X_i\} \cup X_{\pi (i)} |  X_{\pi (i)} }$
 where $X_{\pi (i)}$ is a set of nodes corresponding to 
the 
parent nodes $\pi(i)$ of i.  The underlying distribution represented by a 
 belief network is computed via:%\\
$$BEL  = \bigodot_{i=1}^{n}BEL ^{\downarrow \{X_i\} \cup X_{\pi (i)} |  
X_{\pi (i)} } $$%\\
\end{definition}
The notion of belief network just introduced possesses several important 
characteristics:%\\
\begin{itemize}
\item The relationship between the global belief function $BEL$ and its 
constituting  factors $BEL ^{\downarrow \{X_i\} \cup X_{\pi (i)} |  
X_{\pi (i)} } $ is local: this means that having an empirical model of $BEL$ 
 and knowing the structure $D$ of he distribution $BEL$ we can estimate 
factors $BEL ^{\downarrow \{X_i\} \cup X_{\pi (i)} |  
X_{\pi (i)} } $ separately for each one 
by projection of the empirical model onto the subset $\{X_i\} 
\cup X_{\pi (i)}$ of variables.%\\
\item Removal of a leave node from $D$ (and hence the respective factor from 
the above "product") is equivalent to projection of $BEL$ onto  the space 
spanned by  variables associated with the remaining odes: this enables to 
apply the technique of uncertainty propagation by edge reversals.%\\
\item A belief network reflects causal dependencies among variables (by 
directions of arrows)%\\
\item We can reason about conditional independence of disjunctive sets of 
variables given another set of variables using only graphical properties of 
dag-representation $D$ 
 and without referring to numerical properties of the underlying 
distribution.%\\
\end{itemize} 

In case of probabilistic distributions the first to properties are obvious.
The  last property has been studied very carefully for probabilistic belief 
networks by Geiger, Verma and Pearl \cite{Geiger:90}. It is easily checked 
that the defining formula for the underlying distribution in probabilistic 
case reduces to:%\\
$$P(x_1,...,x_n) = \prod_{i=1}^{n} P(x_i | x_{\pi (i)})$$%\\

In their paper they introduce the notion of d-separation (implying conditional
independence) as follows:

\begin{definition} \cite{Geiger:90}
A {\em trail } in a dag is a sequence of links that form a path in the 
 underlying undirected graph. A node $\beta$ is called a {\em head-to-head 
node}  with 
respect to a trail t if there are two consecutive links $\alpha \rightarrow 
\beta$ and $\beta \leftarrow \gamma$ on that t. %\\
\end{definition}

\begin{definition} \cite{Geiger:90}
A trail t connecting nodes $\alpha$ and $\beta$ is said to be {\em active } 
given a set of nodes L, if (1) every head-to-head-node wrt t either is or has 
a descendent in L and (2) every other node on t is outside L. Otherwise t is 
said to be {\em blocked } (given L).%\\
\end{definition}

\begin{definition} \cite{Geiger:90}
If J,K and L are three disjoint sets of nodes in a dag D, then L is said to 
{\em d-separate } J from K, denoted $I(J,K|L)_D$  iff no trail between a node 
in J and a node in K is active given L.%\\
\end{definition}

It has been shown in \cite{Geiger:90b} that 
\begin{theorem} \cite{Geiger:90b}
Let L be a set of nodes in a dag D, and let $\alpha,\beta \notin L$ be two 
additional nodes in D. Then $\alpha$ and $\beta$ are connected via an active 
trail  (given L) iff  $\alpha$ and $\beta$ are connected via a simple (i.e. 
not possessing cycles in the underlying undirected graph) active trail (given 
L).%\\
\end{theorem}

\begin{definition} \cite{Geiger:90}
If $X_J,X_K,X_L$ are three disjoint sets of variables of a distribution P, 
then $X_J,X_K$ are said to be conditionally independent given $X_L$ (denoted 
$I(X_J,X_K |X_L)_P$ iff $P(x_J,x_K|x_L)=P(x_J|x_L) . P(x_K|x_L)$ for all 
possible values of $X_J,X_K,X_L$ for which $P(x_L)>0$. 
$I(X_J,X_K |X_L)_P$ is called a {\em 
(conditional independence) statement}%\\
\end{definition}

\begin{theorem} \cite{Geiger:90}
Let $P_D=\{P|$(D,P) is a Bayesian network\}. Then:\\

$I(J,K|L)_D$ iff $I(X_J,X_K |X_L)_P$ for all $P \in P_D$.%\\
\end{theorem}

The "only if" part (soundness) states that whenever  $I(J,K|L)_D$ holds in D, 
it must represent an independence that holds in every underlying distribution. 

The "if" part (completeness) asserts that any independence that is not 
detected by d-separation cannot be shared by all distributions in $P_D$ and 
hence cannot be revealed by non-numeric methods. \\

In case of DS belief functions, our defining equation for the underlying 
distribution of a belief network has the form:%\\
$$Bel  = \bigoplus_{i=1}^{n}Bel ^{\downarrow \{X_i\} \cup X_{\pi (i)} |  
X_{\pi (i)} } $$%\\
It is easily seen that the first two properties hold for DS belief networks as 
defined in this paper. The notion of d-separation and its relationship with 
conditional independence are easily transferred onto DS belief networks.%\\
This definition differs significantly from what is generally considered to be 
a DS belief network (e.g. in \cite{Shenoy:90}). Usually, a DS belief network 
is considered to have the defining equation of the form:\\
$$Bel  = \bigoplus_{i=1}^{n}Bel_i$$
where $Bel_i$ is a DS belief function in some subset of variables of $Bel$ 
without any assumption of the nature of $Bel_i$. 
The deviation from traditional 
approach seems to be significant one and requires some explanation. First of 
all the traditional DS belief network has in general none of the 
above-mentioned three  properties hold. The advantages of our DS belief 
network definition are obvious. The question remains whether there are any 
disadvantages. Ones connected with uncertainty propagation within the 
Shenoy/Shafer axiomatic framework \cite{Shenoy:90} is of particular interest 
for implementation of this expert system. \\

Shenoy and Shafer \cite{Shenoy:90} consider it unimportant whether or not the 
factorization 
should refer to conditional probabilities in case of probabilistic belief 
networks. We shall make at this point the remark that for expert system 
inference engine it is of primary importance how the contents of the 
knowledge base should be understood by the  user as beside computation an 
expert system is expected at least to justify its conclusions 
and it can do so only referring to elements of the knowledge base. So if a 
belief network (or a 
hypergraph)
 is to be used as the knowledge base, as much elements as possible 
have to refer to experience of the user.%\\

In our opinion, the major reason for this remark of Shenoy and Shafer is that 
in fact the Dempster-Shafer belief function cannot be decomposed in terms of 
the traditional conditional belief function as defined in the 
previous section 5. 
 This can only be done if a pseudo-conditioning like ours is introduced. 
But an intriguing question remains 
whether the traditional DS belief networks extend essentially the class of 
DS belief functions suitable for Shenoy/Shafer propagation of uncertainty. The 
sad result that really 
\begin{theorem} \cite{Klopotek:93f} 
Traditional DS belief network induced hypergraph (as considered  by
 Shenoy and Shafer \cite{Shenoy:90}) 
may for a given joint belief distribution have simpler structure than
(be properly covered by)
 the closest hypergraph induced by a new DS
belief network (as defined above.)
\end{theorem}
%\\
This fact, however, is compensated completely by another one. Shenoy/Shafer 
propagation does not run in hypergraphs, but in hypertrees. And:\\
\begin{theorem}  \cite{Klopotek:93f} 
No traditional DS belief network induced hypertree  (as considered  by
 Shenoy and Shafer \cite{Shenoy:90}) 
may for a given joint belief distribution have simpler structure than
(be properly covered by)
 the closest hypertree  induced by a new DS
belief network (as defined above.)
\end{theorem}
This fully justifies, in our opinion, the usage of the new belief network 
definition. %\\

%
%อออออออออออออออออออออออออออออออออออออออออออออออออออออออออออออออออออออ
\begin{figure}
\begin{center}
\begin{picture}(8,6)
\put(3,2){\circle*{0.2}} %p7
\put(3,1.3){$p_7$}
\put(3.2,2){\vector(1,0){1.3}}
\put(4.6,2){\line(1,0){1.7}}
\put(6.6,2){\circle*{0.2}}%p8 
\put(6.6,1.3){$p_8$}
\put(3,5.6){\circle*{0.2}}%p4
\put(3,6.1){$p_4$} 
\put(3.2,5.4){\vector(1,-1){0.8}}    %skosna
\put(4.1,4.5){\line(1,-1){0.4}}    %skosna
\put(3.2,5.6){\vector(1,0){1.3}}
\put(4.6,5.6){\line(1,0){1.7}}
\put(6.6,5.6){\circle*{0.2}}%p5 
\put(6.6,6.1){$p_5$}
\put(6.6,5.4){\vector(0,-1){1.6}}
\put(6.6,3.4){\line(0,-1){1.2}}
\put(1.2,5.6){\circle*{0.2}}%p3 
\put(1.2,6.1){$p_3$}
\put(1.4,5.6){\vector(1,0){0.9}}
\put(2.4,5.6){\line(1,0){0.4}}
\put(1.2,3.8){\circle*{0.2}}%p1 
\put(1.2,4.3){$p_1$}
\put(1.4,3.8){\vector(1,0){0.9}}
\put(2.4,3.8){\line(1,0){0.4}}
\put(3,3.8){\circle*{0.2}} %p2
\put(2.4,4.3){$p_2$}
\put(3,4){\vector(0,1){0.9}}
\put(3,5){\line(0,1){0.4}}
\put(3,3.6){\vector(0,-1){0.9}}
\put(3,2.6){\line(0,-1){0.4}}
\put(4.7,3.8){\circle*{0.2}} %p6
\put(5,4.1){$p_6$}
\put(4.9,3.6){\vector(1,-1){1.3}}    %skosna
\end{picture}
\end{center}
\caption{Belief Network - An Example}  \label{ryssiec}
\end{figure}
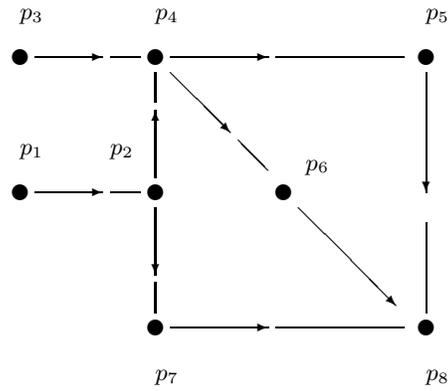%\\
%อออออออออออออออออออออออออออออออออออออออออออออออออออออออออออออออออออออ

\subsection{Understanding a Belief Network in Terms of Rules}

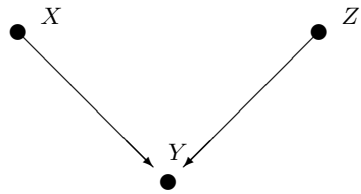
\begin{figure} 
\begin{center}
\begin{picture}(5.5,3.5)(0,0)
% Die Knoten 
\put(1.3,3.1){$X$}
\put(3.0,1.3){$Y$}
\put(5.3,3.1){$Z$}
\put(1.0,3.0){\circle*{0.2}} 
\put(3.0,1.0){\circle*{0.2}} 
\put(5.0,3.0){\circle*{0.2}} 
% Die Verbindungspfeile
\put(1.0,3.0){\vector(1,-1){1.8}} % XY
\put(5.0,3.0){\vector(-1,-1){1.8}} % ZY
\end{picture}%\\
\end{center}
\caption{A Typical Node of a Belief Network}  \label{ryswezel}
\end{figure}

A dag structure of a belief network was presented in Fig.\ref{ryssiec}. 
Let us consider a fragment of a belief network in Fig.\ref{ryswezel}
consisting of a node (and the associated variable, or attribute, or feature)  
Y and all of its parents: X and  Z. Let the values of them range 
for Y: $\{y_1,y_2,y_3\}$, for X:$\{x_1,x_2\}$, for 
Z:$\{z_1,z_2,z_3,z_4\}$,  resp.\\

%อออออออออออออออออออออออออออออออออออออออออออออออออออออออออออออออออออออ

%\\
%อออออออออออออออออออออออออออออออออออออออออออออออออออออออออออออออออออออ

Then a table of valuations will be associated with the node Y.
In probabilistic case the valuation will take the form given in table          
\ref{tabwp}.%\\

\begin{table}
\caption{Probabilistic Node Valuation} \label{tabwp}
\begin{center}
\begin{tabular}{lllr}
\hline
X & Z & Y & Cond. Probability $P(y | x,z)$\\
\hline
$x_1$ & $z_1$ & $y_1$ & $p_{111}$\\
$x_1$ & $z_1$ & $y_2$ & $p_{112}$\\
$x_1$ & $z_1$ & $y_3$ & $p_{113}$\\
$x_1$ & $z_2$ & $y_1$ & $p_{121}$\\
$x_1$ & $z_2$ & $y_2$ & $p_{122}$\\
$x_1$ & $z_2$ & $y_3$ & $p_{123}$\\
$x_1$ & $z_3$ & $y_1$ & $p_{131}$\\
$x_1$ & $z_3$ & $y_2$ & $p_{132}$\\
$x_1$ & $z_3$ & $y_3$ & $p_{133}$\\
$x_1$ & $z_4$ & $y_1$ & $p_{141}$\\
$x_1$ & $z_4$ & $y_2$ & $p_{142}$\\
$x_1$ & $z_4$ & $y_3$ & $p_{143}$\\
$x_2$ & $z_1$ & $y_1$ & $p_{211}$\\
$x_2$ & $z_1$ & $y_2$ & $p_{212}$\\
$x_2$ & $z_1$ & $y_3$ & $p_{213}$\\
$x_2$ & $z_2$ & $y_1$ & $p_{221}$\\
$x_2$ & $z_2$ & $y_2$ & $p_{222}$\\
$x_2$ & $z_2$ & $y_3$ & $p_{223}$\\
$x_2$ & $z_3$ & $y_1$ & $p_{231}$\\
$x_2$ & $z_3$ & $y_2$ & $p_{232}$\\
$x_2$ & $z_3$ & $y_3$ & $p_{233}$\\
$x_2$ & $z_4$ & $y_1$ & $p_{241}$\\
$x_2$ & $z_4$ & $y_2$ & $p_{242}$\\
$x_2$ & $z_4$ & $y_3$ & $p_{243}$\\
\hline
\end{tabular}
\end{center}
\end{table}

In general, it will be an exhaustive table of conditional probabilities
of each value of a node given each value combination of its parents. 
In case of Dempster-Shafer valuation, only some (that is focal point) 
conditional valuations will be given, e.g. as in table \ref{tabwds}.%\\

\begin{table}
\caption{Dempster-Shafer Valuation of a Node} \label{tabwds}
\begin{center}
\begin{tabular}{lllr}
\hline
X & Z & Y & bpa function $m ^{\downarrow \{Y,X,Z\} | \{X, Z\} }(y,x,z)$\\
\hline
$\{(x_1,$ & $z_1,$ & $y_1)\}$ & $m_1 $\\
$\{(x_1,$ & $z_1,$ & $y_2), $ &       \\
$  (x_1,$ & $z_3,$ & $y_2)\}$ & $m_2 $\\
$\{(x_2,$ & $z_2,$ & $y_3)\}$ & $m_3 $\\
$\{(x_1,$ & $z_4,$ & $y_1), $ &       \\
$  (x_1,$ & $z_4,$ & $y_3), $ &       \\
$  (x_2,$ & $z_4,$ & $y_1), $ &       \\
$  (x_2,$ & $z_4,$ & $y_3)\}$ & $m_4 $\\
\hline
\end{tabular}
\end{center}
\end{table}

Each table of this form may be displayed for presentation purposes as a "beam" 
of rules: \\
 in probabilistic case:\\

\noindent
{\bf IF }  X = $x_1$ {\bf AND } Z = $z_1$ {\bf THEN }  Y = $y_1$ {\bf WITH }
 $p_{111}$\\
{\bf IF }  X = $x_1$ {\bf AND } Z = $z_1$ {\bf THEN }  Y = $y_2$ {\bf WITH }
 $p_{112}$\\
{\bf IF }  X = $x_1$ {\bf AND } Z = $z_1$ {\bf THEN }  Y = $y_3$ {\bf WITH }
 $p_{113}$\\
{\bf IF }  X = $x_2$ {\bf AND } Z = $z_1$ {\bf THEN }  Y = $y_1$ {\bf WITH }
 $p_{211}$\\
\dots \dots \dots\\
{\bf IF }  X = $x_2$ {\bf AND } Z = $z_4$ {\bf THEN }  Y = $y_3$ {\bf WITH }
 $p_{243}$\\
%\\
and in Dempster-Shafer case for the valuation in table \ref{tabwds}:\\
\noindent
{\bf IF }
X=$x_1$  {\bf AND } Z=$z_1$  {\bf THEN }  Y=$y_1$ 
{\bf WITH }  $Q_1$\\
{\bf IF }
X=$x_1$  {\bf AND }  Z=$z_1$   {\bf THEN }  Y=$y_2$  \\
{\bf AND IF }
X=$x_1$  {\bf AND }  Z=$z_3$   {\bf THEN }  Y=$y_2$ 
{\bf WITH }  $Q_2$\\
{\bf IF }
X=$x_2$  {\bf AND } Z=$z_2$  {\bf THEN }  Y=$y_3$ 
{\bf WITH }  $Q_3$\\
{\bf IF }
X=$x_1$   {\bf AND }  Z=$z_4$  {\bf THEN }  Y=$y_1$ \\
{\bf AND IF }
X=$x_1$   {\bf AND }  Z=$z_4$  {\bf THEN }  Y=$y_3$ \\
{\bf AND IF }
  X=$x_2$  {\bf AND }  Z=$z_4$  {\bf THEN }  Y=$y_1$ \\
{\bf AND IF }
  X=$x_2$  {\bf AND }  Z=$z_4$  {\bf THEN }  Y=$y_3$ \\
{\bf WITH }  $Q_4$\\

while measures $Q$ (commonality)  are calculated according to the principles 
of the Dempster-Shafer Theory     \cite{Shafer:76}.%\\

First of all notice that we are talking always about a "beam of rules"
and not of a single rule -  this was rarely considered in production 
rule systems.%\\

In probabilistic case, it should be clear from the very beginning what is 
meant by a single rule (of the beam): the certainty factor after the word 
"with"  expresses the proportion of cases fitting the 
"rule" among those meeting the premise. At the same time, the above examples 
exhaust the richness of the language needed for presenting contents of the 
knowledge base to the user. \\

The Dempster-Shafer case is more subtle. Here, all the variables (X,Y,Z) are 
 treated as potentially set-valued, and the equality sign in the 
expressions should be rather treated as "is containing" and not as set 
equality sign.  The certainty factor after the word 
"with"  expresses again the proportion of cases fitting the 
"rule" among those meeting the premise. At the same time, the above examples 
exhaust the richness of the language needed for presenting contents of the 
knowledge base to the user. \\

The representation language for knowledge base has not only syntactic 
restrictions, but also semantica ones: within a rule, only direct predecessors 
of a given node may occur in the premise part of a rule.\\

\subsection{Rule-oriented Queries to the Knowledge Base} 

A knowledge base in belief network notation means something more than just the 

sum of rules. The built-in system of uncertainty propagation can for the given 
network give the level of probability (belief) for more general formulae
without any restrictions provided by network topology:\\
\begin{itemize}%\\
\item for logical expressions (with operators AND, OR, NOT)  
in atomic formulae of the form   $<$variable$>$ = $<$value$>$. \\
- the system may deduce what is the unconditional probability/belief of such a 
formulae to hold,\\
- the same question may be also answered conditionally: 
what is the probability/belief of the expression given some of the variables
of the network have restricted value ranges,\\
- the same question may be also answered conditionally: 
what is the probability/belief of the expression given some constraints 
expressed in terms of
logical expressions are imposed 
onto  the network,\\
\item for beams of rules of the type IF attr1=? AND attr2=? AND 
attr3=? ... THEN attr=? %\\
- the complete beam can be inferred unconditionally, or conditionally or for a 
given set of constraints as described for logical expressions.\\
\end{itemize}%\\

For probabilistic case, the methods of implementation of the above  
query-answering is described in  \cite{Pearl:1}. Implementation for 
Dempster-Shafer Theory runs along the same lines. Let us consider as an 
example the network in Fig.\ref{ryssiec}. Let us assume that each of the nodes 
$p_1$,...,$p_8$ takes only one of two values: 'y' and 'n'. %\\

\begin{figure}
\begin{center}
\begin{picture}(8,6)

% Anfrage x1 : p1 p5
\put(7.7,5.6){\circle{0.2}} %x1
\put(7.7,6.1){$x_1$}
% von p1: 
\put(1.4,3.9){\vector(4,1){6.1}}
% von p5
\put(6.6,5.6){\vector(1,0){0.9}} 

\put(3,2){\circle*{0.2}} %p7
\put(3,1.3){$p_7$}
\put(3.2,2){\vector(1,0){1.3}}
\put(4.6,2){\line(1,0){1.7}}

\put(6.6,2){\circle*{0.2}}%p8 
\put(6.6,1.3){$p_8$}

\put(3,5.6){\circle*{0.2}}%p4
\put(3,6.1){$p_4$} 
\put(3.2,5.4){\vector(1,-1){0.8}}    %skosna
\put(4.1,4.5){\line(1,-1){0.4}}    %skosna
\put(3.2,5.6){\vector(1,0){1.3}}
\put(4.6,5.6){\line(1,0){1.7}}
\put(6.6,5.6){\circle*{0.2}}%p5 
\put(6.6,6.1){$p_5$}
\put(6.6,5.4){\vector(0,-1){1.6}}
\put(6.6,3.4){\line(0,-1){1.2}}
\put(1.2,5.6){\circle*{0.2}}%p3 
\put(1.2,6.1){$p_3$}
\put(1.4,5.6){\vector(1,0){0.9}}
\put(2.4,5.6){\line(1,0){0.4}}
\put(1.2,3.8){\circle*{0.2}}%p1 
\put(1.2,4.3){$p_1$}
\put(1.4,3.8){\vector(1,0){0.9}}
\put(2.4,3.8){\line(1,0){0.4}}
\put(3,3.8){\circle*{0.2}} %p2
\put(2.4,4.3){$p_2$}
\put(3,4){\vector(0,1){0.9}}
\put(3,5){\line(0,1){0.4}}
\put(3,3.6){\vector(0,-1){0.9}}
\put(3,2.6){\line(0,-1){0.4}}
\put(4.7,3.8){\circle*{0.2}} %p6
\put(5,4.1){$p_6$}
\put(4.9,3.6){\vector(1,-1){1.3}}    %skosna
\end{picture}
\end{center}
\caption{A Belief Network with Query-Node: $x_1: 
p_5='y' .or. p_1='n' $
}  \label{ryssiec2}
\end{figure}
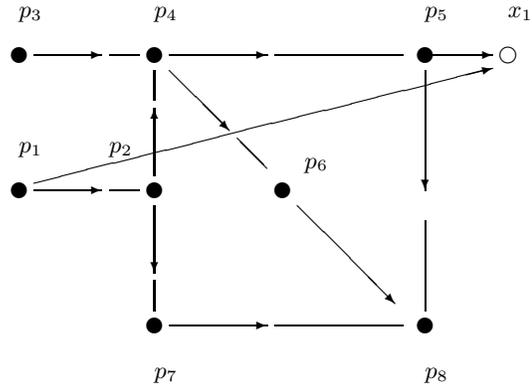

We intend to ask the question: 
$p_5='y' .or. p_1='n' $. The belief network is to be amended temporarily by a 
node  $x_1$, taking values 
'y' i 'n', which we connect with $p_5$  and $p_1$ (see 
Fig.\ref{ryssiec2}). With this node, we associate the rules:\\
{\bf IF }$p_5='t'$ {\bf AND }$p_1='n'$ {\bf THEN }$x_1='t'${\bf WITH } $1.$\\
{\bf IF }$p_5='t'$ {\bf AND }$p_1='t'$ {\bf THEN }$x_1='t'${\bf WITH } $1.$\\
{\bf IF }$p_5='n'$ {\bf AND }$p_1='n'$ {\bf THEN }$x_1='t'${\bf WITH } $1.$\\
{\bf IF }$p_5='n'$ {\bf AND }$p_1='t'$ {\bf THEN }$x_1='n'${\bf WITH } $1.$\\
and for the remaining rules the probability is equal 0.\\
%อออออออออออออออออออออออออออออออออออออออออออออออออออออออออออออออออออออ
%\\
%อออออออออออออออออออออออออออออออออออออออออออออออออออออออออออออออออออออ\\
The answer is the final valuation of the node $x_1$ 
after uncertainty propagation.\\

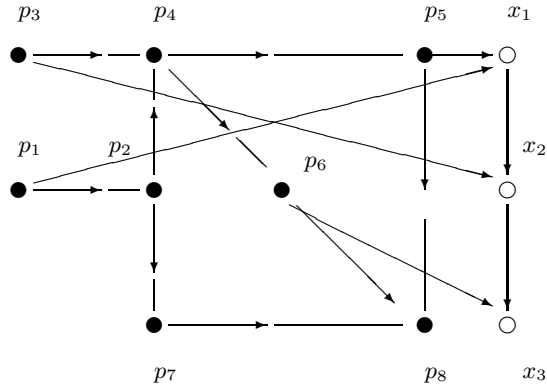
\begin{figure}
\begin{center}
\begin{picture}(8,6)

% Anfrage x1 : p1 or  p5
\put(7.7,5.6){\circle{0.2}} %x1
\put(7.7,6.1){$x_1$}
% von p1: 
\put(1.4,3.9){\vector(4,1){6.1}}
% von p5
\put(6.6,5.6){\vector(1,0){0.9}} 

% Anfrage x2 : x1 and  p3 
\put(7.7,3.8){\circle{0.2}} %x2
\put(7.9,4.3){$x_2$}
% von x1
\put(7.7,5.4){\vector(0,-1){1.4}} 
% von p3
\put(1.4,5.5){\vector(4,-1){6.1}}

% Anfrage x3 : if x2 then p6
\put(7.7,2.0){\circle{0.2}} %x3
\put(7.9,1.3){$x_3$}
% von x2
\put(7.7,3.6){\vector(0,-1){1.4}} 
% von p6
\put(4.8,3.6){\vector(2,-1){2.7}}    %skosna

\put(3,2){\circle*{0.2}} %p7
\put(3,1.3){$p_7$}
\put(3.2,2){\vector(1,0){1.3}}
\put(4.6,2){\line(1,0){1.7}}

\put(6.6,2){\circle*{0.2}}%p8 
\put(6.6,1.3){$p_8$}

\put(3,5.6){\circle*{0.2}}%p4
\put(3,6.1){$p_4$} 
\put(3.2,5.4){\vector(1,-1){0.8}}    %skosna
\put(4.1,4.5){\line(1,-1){0.4}}    %skosna
\put(3.2,5.6){\vector(1,0){1.3}}
\put(4.6,5.6){\line(1,0){1.7}}
\put(6.6,5.6){\circle*{0.2}}%p5 
\put(6.6,6.1){$p_5$}
\put(6.6,5.4){\vector(0,-1){1.6}}
\put(6.6,3.4){\line(0,-1){1.2}}
\put(1.2,5.6){\circle*{0.2}}%p3 
\put(1.2,6.1){$p_3$}
\put(1.4,5.6){\vector(1,0){0.9}}
\put(2.4,5.6){\line(1,0){0.4}}
\put(1.2,3.8){\circle*{0.2}}%p1 
\put(1.2,4.3){$p_1$}
\put(1.4,3.8){\vector(1,0){0.9}}
\put(2.4,3.8){\line(1,0){0.4}}
\put(3,3.8){\circle*{0.2}} %p2
\put(2.4,4.3){$p_2$}
\put(3,4){\vector(0,1){0.9}}
\put(3,5){\line(0,1){0.4}}
\put(3,3.6){\vector(0,-1){0.9}}
\put(3,2.6){\line(0,-1){0.4}}
\put(4.7,3.8){\circle*{0.2}} %p6
\put(5,4.1){$p_6$}
\put(4.9,3.6){\vector(1,-1){1.3}}    %skosna
\end{picture}
\end{center}
\caption{%
 $x_3: \ if \ (p_5='t' \ .or. \  p_1='n') \ .and.\  p_3='n' 
\ then \ p6='t' $
}  \label{ryssiec3}
\end{figure}

Now let's turn to checking validity of the query of the form: %\\
$$x_3: \ if \ (p_5='t' \ .or. \  p_1='n') \ .and.\  p_3='n' 
\ then \ p6='t' $$ 
with respect to contents of the knowledge base.
Node  $x_1$ connected with  $p_5$ and $p_1$, 
node  $x_2$ connected with $p_3$ and $p_1$ and 
node  $x_3$  connected with $p_6$ and $p_1$ (see   Fig.\ref{ryssiec3}). 
Let nodes   $x_1$ and $x_2$ take values 'y' or 'n', and the node
$x_3$ 
- 'y', 'n', '?'. Node     $x_1$ is associated with a 
beam of rules as mentioned 
above, and  $x_2$ - with:\\
{\bf IF }$p_3='t'$ {\bf AND }$x_1='t'$ {\bf THEN }$x_2='t'${\bf WITH } $1.$\\
{\bf IF }$p_3='t'$ {\bf AND }$x_1='n'$ {\bf THEN }$x_2='n'${\bf WITH } $1.$\\
{\bf IF }$p_3='n'$ {\bf AND }$x_1='t'$ {\bf THEN }$x_2='n'${\bf WITH } $1.$\\
{\bf IF }$p_3='n'$ {\bf AND }$x_1='n'$ {\bf THEN }$x_2='n'${\bf WITH } $1.$\\
with the remaining rule probabilities equal 0.\\
Node     $x_3$ is associated with rule beam:\\
{\bf IF }$p_6='t'$ {\bf AND }$x_2='t'$ {\bf THEN }$x_3='t'${\bf WITH } $1.$\\
{\bf IF }$p_6='t'$ {\bf AND }$x_2='n'$ {\bf THEN }$x_3='n'${\bf WITH } $1.$\\
{\bf IF }$p_6='n'$ {\bf AND }$x_2='t'$ {\bf THEN }$x_3='?'${\bf WITH } $1.$\\
{\bf IF }$p_6='n'$ {\bf AND }$x_2='n'$ {\bf THEN }$x_3='?'${\bf WITH } $1.$\\
with the remaining rule probabilities equal 0.\\
%อออออออออออออออออออออออออออออออออออออออออออออออออออออออออออออออออออออ\\
 %\\

%อออออออออออออออออออออออออออออออออออออออออออออออออออออออออออออออออออออ\\
The final answer is the valuation of the node  $x_3$ 
after uncertainty propagation. The probability of $x_3='t'$ indicates how
frequently the rule fires and is correct,%\\
probability of the event $x_3='?'$ 
indicates how often the rule did not fire (and only for this reason is 
considered by logicians to be true).\\
probability of  $x_3='n'$ 
indicates, how frequently the rule fires, but is in error.
(The respective 3-valued logic is considered in \cite{Klopotek:Wien92}.
%\\

\subsection{Belief Network Construction}

To construct a belief network correctly, one should:
\begin{itemize}
\item properly uncover the causal structure governing the attribute, 
and thereafter
\item calculate the valuations of each of the node of the network.
\end{itemize}

For the probabilistic case, a number of respective causal-structure-from-data
reconstruction algorithms have been elaborated
\cite{Acid:1},
\cite{Chow:1},
\cite{Chow:2},
\cite{Cooper:92},
\cite{Geiger:90b},
\cite{Rebane:1},
\cite{Spirtes:90b},
\cite{Spirtes:91},
\cite{Srinivas:1},
\cite{Valiveti:92},
\cite{Verna:90},
\cite{Wermuth:83}.

Node valuation can be also calculated from data \cite{Chow:2}
as relative conditional frequency. 

The Dempster-Shafer case is more difficult as clear relationship between
data and the joint belief distribution is still subject of disputes.
Set this issue aside, proper Q-value quotients may be exploited.
 %\\

However, the knowledge base may be only partially reflected by data. Thenm 
there emerges the necessity to manipulate manually the knowledge base.
The user is then requested to enter a consistent description of the network 
structure. Given this, the entrance of the valuations may be run precisely 
following the vary same pattern as the contents of the knowledge base are 
presented to the user.\\

   {}

   \end{document}